\documentclass{article}
\usepackage{ijcai24}

\usepackage{times}
\usepackage{soul}
\usepackage{url}
\usepackage[hidelinks]{hyperref}
\usepackage[utf8]{inputenc}
\usepackage[small]{caption}
\usepackage{graphicx}
\usepackage{amsmath}
\usepackage{amsthm}
\usepackage{booktabs}
\usepackage[ruled,vlined,linesnumbered]{algorithm2e}
\usepackage{algorithmic}
\usepackage[switch]{lineno}
\usepackage{color}
\usepackage{enumitem}
\usepackage{subfigure}

\title{WESE: Weak Exploration to Strong Exploitation for LLM Agents}

\author{
    Anonymous Authors
    \affiliations
    Anonymous Affiliation
    \emails
    anonymous@example.com
}

\author{
Xu Huang$^1$
\and
Weiwen Liu$^2$\and
Xiaolong Chen$^1$\and
Xingmei Wang$^1$ \and \\
Defu Lian$^{1}$\footnote{Defu Lian is the corresponding author.} \and
Yasheng Wang$^2$ \and
Ruiming Tang$^2$ \and
Enhong Chen$^1$
\affiliations
$^1$University of Science and Technology of China, Hefei, China\\
$^2$Huawei Noah’s Ark Lab, Shenzhen, China
\emails
{xuhuangcs, chenxiaolong, xingmeiwang}@mail.ustc.edu.cn,
\{liandefu, cheneh\}@ustc.edu.cn,\\
\{liuweiwen8,wangyasheng, tangruiming\}@huawei.com
}
\begin{document}

\maketitle

\begin{abstract}
Recently, large language models (LLMs) have demonstrated remarkable potential as an intelligent agent. However, existing researches mainly focus on enhancing the agent's reasoning or decision-making abilities through well-designed prompt engineering or task-specific fine-tuning, ignoring the procedure of exploration and exploitation. When addressing complex tasks within open-world interactive environments, these methods exhibit limitations. Firstly, the lack of global information of environments leads to greedy decisions, resulting in sub-optimal solutions. On the other hand, irrelevant information acquired from the environment not only adversely introduces noise, but also incurs additional cost.
This paper proposes a novel approach, \textbf{W}eak \textbf{E}xploration to \textbf{S}trong \textbf{E}xploitation (\textbf{WESE}), to enhance LLM agents in solving open-world interactive tasks. Concretely, WESE involves decoupling the exploration and exploitation process, employing a cost-effective weak agent to perform exploration tasks for global knowledge. A knowledge graph-based strategy is then introduced to store the acquired knowledge and extract task-relevant knowledge, enhancing the stronger agent in success rate and efficiency for the exploitation task. Our approach is flexible enough to incorporate diverse tasks, and obtains significant improvements in both success rates and efficiency across four interactive benchmarks.

\end{abstract}

\section{Introduction}

Large language models (LLMs) showcase a myriad of capabilities across diverse domains, encompassing human-computer conversation, instruction following, reasoning, and few-shot learning~\cite{zhao2023survey}. These comprehensive abilities form a robust foundation, positioning LLMs as intelligent agents in solving open-world tasks, such as household tasks and open-world question-answering tasks~\cite{wang2023survey,xi2023rise}. Recently, there have been numerous works to investigate the potential of LLM agents in enhancing their capabilities for open-world tasks.

Benefiting from the capabilities of LLMs in instruction-following and few-shot learning, most methods guide LLMs in decision-making tasks through human-crafted design, avoiding the costly fine-tuning of LLMs~\cite{wei2022chain,wang2022self,yao2022react,kojima2022large}. Existing prompt-engineering approaches primarily consider two factors: how to incorporate task-relevant information in the prompt, and how to elicit the reasoning ability of LLMs through prompts. Task-relevant information encompasses task descriptions and contextual feedback, such as the question and pertinent task statements in question-answering tasks, along with textual materials retrieved by the agent from the web while problem-solving. To enhance the reasoning capabilities of LLM agents, methods like CoT~\cite{wei2022chain}, ReAct~\cite{yao2022react} Reflexion~\cite{shinn2023reflexion}, et al, inspire LLMs to engage in reasoning by constructing few-shot examples with explicit reasoning paths. 
% However, due to the challenging characteristics of open-world tasks, such methods exhibit some limitations.

\begin{figure}[t]
    \centering
    \subfigure[ScienceWorld. Lack of global environmental information causes failure due to trapping in a loop or sub-optimal solution.]{
        \includegraphics[width=0.9\columnwidth]{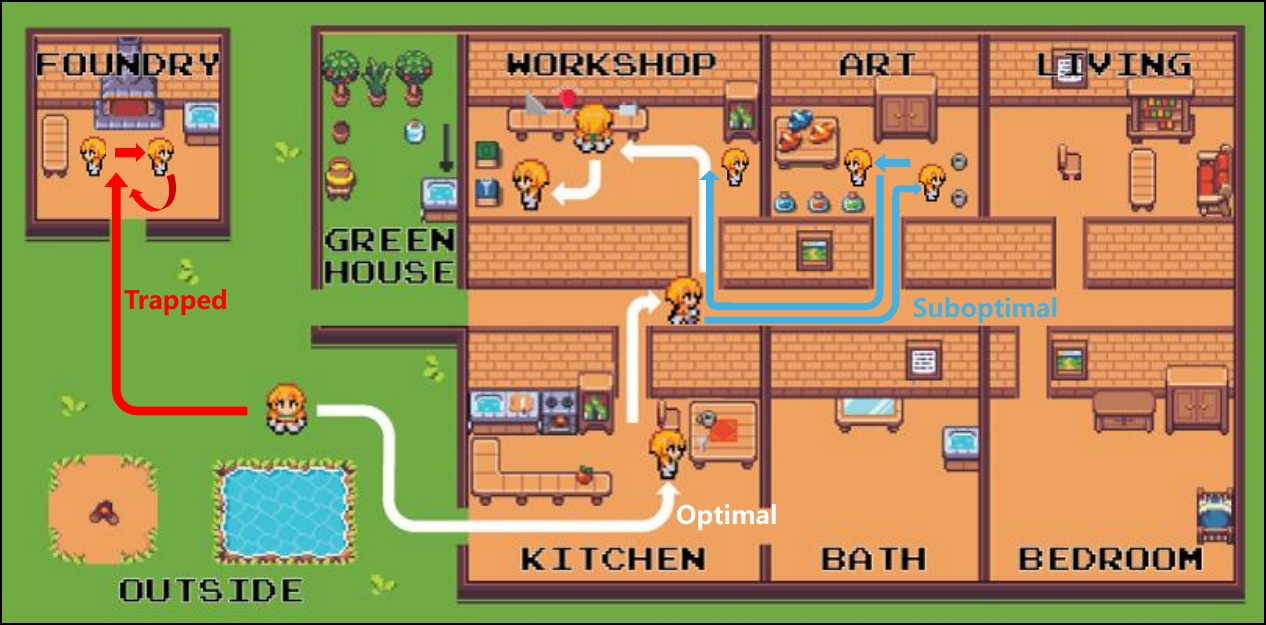}
        \label{fig:suboptimal_sol}
    }
    \subfigure[HotPotQA. The green sentence is helpful while others are task-irrelevant.]{
        \includegraphics[width=0.9\columnwidth]{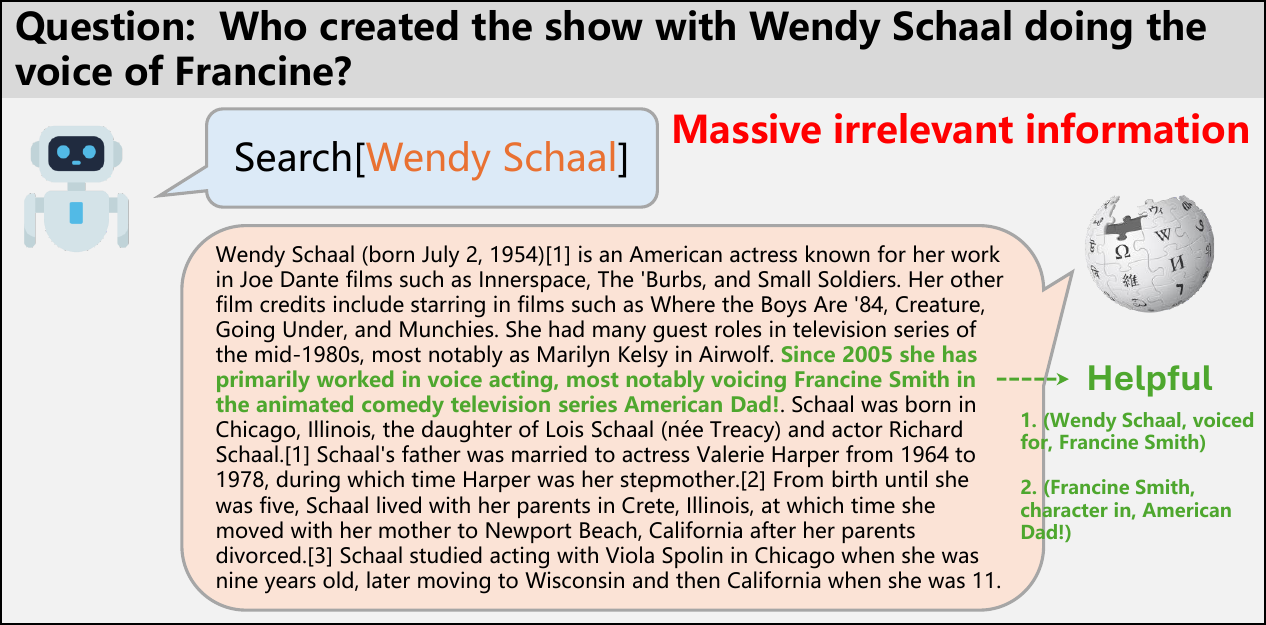}
        \label{fig:irrelevant_info}
    }
    \vspace{-0.3cm}
    \caption{Examples for sub-optimal decisions and irrelevant information in feedbacks.}
\end{figure}

However, open-world tasks serve as a simulation of the real environment, wherein an agent explores and interacts continuously with the environment to acquire more information for solving complex tasks~\cite{cote2019textworld,shridhar2020alfworld,wang2022scienceworld}. There are several characteristics of such tasks, making them more challenging. The ability of LLM agents is far from optimal due to the following challenges: 1) \textbf{Complexity}. Each task involves multi-step actions and each task can have multiple feasible solutions. 2) \textbf{Uncertainty}. The agent cannot obtain all the information from the initial task description, and additional information must be acquired through exploration. 
Regarding these challenges, solving these tasks necessitates multi-step exploration and exploitation by the agent. Exploration involves perceiving the environment and obtaining task-relevant information, while exploitation involves making action decisions based on existing knowledge. In existing prompt-based methods, exploration and exploitation issues are often overlooked, embedded within the reasoning process of the LLM~\cite{yao2022react}, leading to two major problems. 

Firstly,\textit{\textbf{ the lack of global awareness of the environment at the outset solutions results in suboptimal decision-making by the LLM}}. As illustrated in Figure~\ref{fig:suboptimal_sol}, the goal is to find one aluminum object and test its conductivity. The agent is located outside initially. The best trajectory is marked with the white line, where the agent goes to the kitchen to take the aluminum fork first and then go to the workshop. When lack of global environmental information, the agent probably gets trapped in some room due to failure in finding an aluminum object (the red line) or chooses a more time-consuming way (the blue line). 
Secondly, \textbf{\textit{the knowledge acquired by the LLM from environmental exploration tends to be excessive, including irrelevant information to the task.}} The presence of such information not only disrupts LLM decision-making but also incurs additional costs. Referred in Figure~\ref{fig:irrelevant_info}, the feedback from the environment usually consists of massive task-irrelevant information while only one helpful sentence, i.e. the green line in this example, resulting in extra token usage of LLM and a negative effect on making optimal decisions.

To address the above limitations, we propose a novel prompt-based strategy to enhance the LLM agent in this work, termed \textbf{W}eak \textbf{E}xploration to \textbf{S}trong \textbf{E}xploitation (\textbf{WESE}). To tackle the first limitation, we introduce an idea that decouples the exploration and exploitation. Specifically, we construct two distinct LLM agents for exploration and exploitation tasks, respectively. In the exploration task, the LLM agent's goal is to interact with the environment, exploring potentially helpful environmental information for task resolution. In the exploitation task, the information obtained during exploration serves as a global environmental prior, aiding the LLM agent in reasoning and decision-making to generate decisions.
Regarding the second limitation, we compress the environmental information acquired by the exploration agent, structuring it in the form of a knowledge graph. During exploitation, we adopt a one-hop knowledge retrieval approach, selecting one-hop neighbors of task-relevant entities from the graph as priors, thereby reducing interference from irrelevant information. Furthermore, to further minimize resource consumption, we observe that a cost-effective weaker LLM (such as a 7B model) is fully capable of the less challenging exploratory tasks. Therefore, we propose the strategy of \textit{weak exploration to strong exploitation}---leveraging the knowledge explored by the weak LLM agent to enhance the performance of the strong LLM agent.

Our main contributions are summarized as follows: 
\begin{itemize}[leftmargin=*]
    \item To the best of our knowledge, this is the first work to investigate the effect of decoupling exploration and exploitation for LLM agents in open-world tasks. We further propose WESE, leveraging a weaker agent to enhance the stronger agent in a cost-effective manner.
    \item To better leverage the environmental information obtained from exploration, we introduce a strategy to compress it into a knowledge graph. Then we devise a one-hop retrieval approach to filter out the irrelevant information.
    \item Experimental results over four open-world interactive benchmarks demonstrate the superiority of WESE, notably in achieving a remarkable balance between effectiveness, efficiency and cost.
\end{itemize}
\section{Related Works}
\subsection{LLM agents}
With the emergence of LLMs, their intelligence has sparked considerable potential in applying LLMs as the brains of agents.
Existing LLM agent works primarily consider three key modules: planning, tool usage, and memory~\cite{wang2023survey}. Planning module aims to empower agent with the task-decomposition ability, encompassing works on task decomposition~\cite{wang2023plan}, feedback-driven adjustments~\cite{shinn2023reflexion}, and multi-path reasoning~\cite{yao2023tree,besta2023graph}. 
Tool usage aims to strengthen the ability to use external tools~\cite{qin2023tool}. For instance, Visual ChatGPT~\cite{wu2023visual} incorporates visual models as tools to augment the LLM's visual capabilities. ToolLlama~\cite{qin2023toolllm} fine-tunes Llama's ability to leverage various APIs. The memory module focuses on storing feedback information perceived from the environment, assisting the agent with experience, and fostering the growth of the agent. In Generative Agents~\cite{park2023generative}, memories of simulated roles are stored as texts, utilizing RAG for relevant pieces. REMEMBER~\cite{zhang2023large} proposes a semi-parametric memory, i.e. the Q-value table, to record rewards as the value and action in a given environment and task as the key. MemoryBank~\cite{wang2023augmenting} leverages the Ebbinghaus forgetting curve, incorporating update and forgetting mechanisms into the memory design.

In our proposed WESE, the knowledge graph is essentially a memory, updating information obtained through exploration into the graph.

\subsection{LLM for open-world tasks}
Open-world tasks represent the simulation of real-world environments. Within these tasks, agents engage in continuous interactions with the environment to gather pertinent information, subsequently making decisions and taking action to accomplish goals. Open-world tasks typically exhibit fewer constraints on the process, placing greater emphasis on the final rewards. Representative examples of open-world tasks include games like ``Minecraft"~\cite{wang2023voyager,wang2023describe}, where textual information and visual feedback are involved. Another category comprises text-based simulators based on the TextWorld~\cite{cote2019textworld}, such as AlfWorld~\cite{shridhar2020alfworld}, which involves household tasks, ScienceWorld~\cite{wang2022scienceworld}, which involves simple scientific experiments, and question-answering tasks~\cite{yang2018hotpotqa,thorne2018fever} where agents need to interact with the web to obtain supporting information, such as Wikipedia.
In tackling such tasks, Chain-of-Thought(CoT)~\cite{wei2022chain} proposes adding few-shot examples in the prompt, guiding the LLM to solve the task step by step. ReAct~\cite{yao2022react} induces the reasoning capability of LLMs by introducing an extra thought step. Subsequent methods have built upon ReAct, with enhancements such as the Reflexion~\cite{shinn2023reflexion} mechanism, allowing agents to learn from mistakes in subsequent attempts. Additionally, several methods leverage the coding capabilities of LLMs, transforming tasks into programming tasks and guiding LLMs to generate codes as plans, such as VOYAGER~\cite{wang2023voyager}.

\section{Methodologies}
% We introduce our method WESE in this section. The framework and the pseudo-code are illustrated in Figure~\ref{fig:framework} and Algorithm~\ref{alg:wese}, respectively. 

\begin{figure*}[t]
    \centering
    \includegraphics[width=0.90\textwidth]{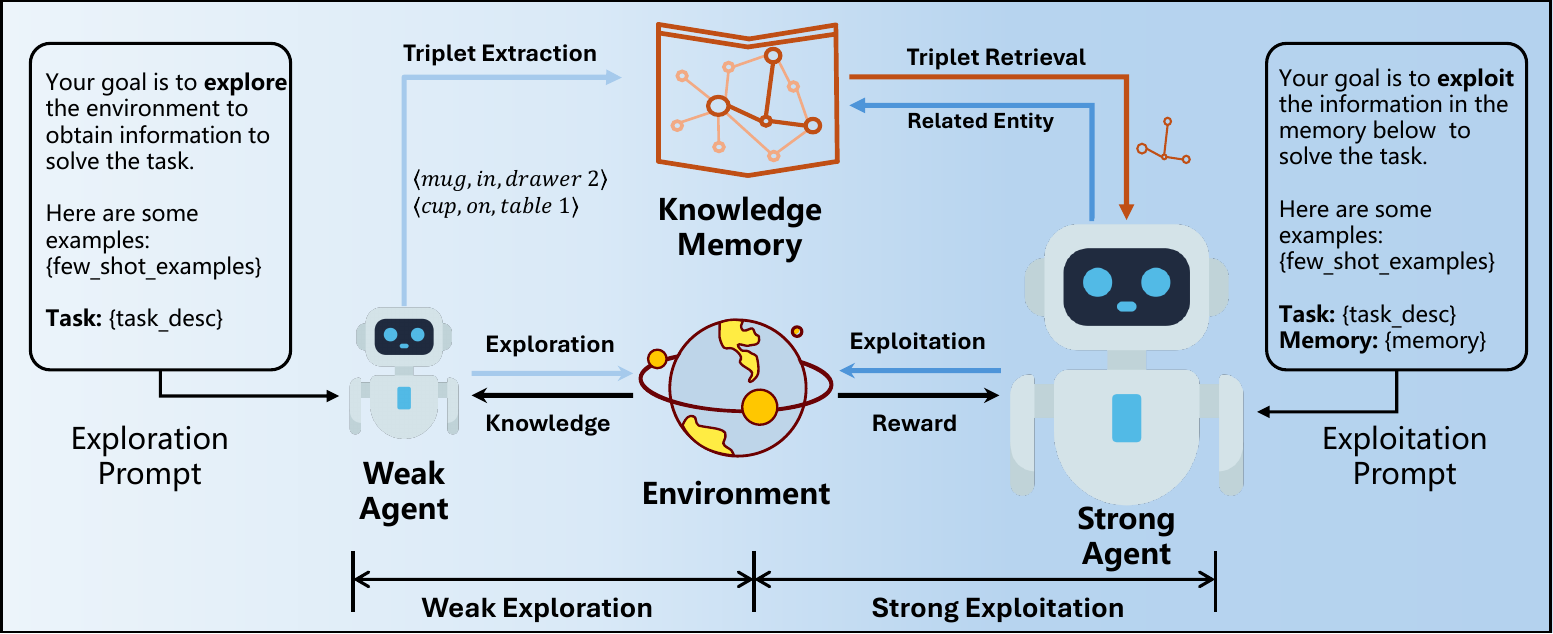}
    \caption{Framework of WESE. The left part represents the weak exploration and the right part represents the strong exploitation. We employ Llama-2-7B as the weak agent and text-davinci-003 as the strong agent in the implementation.}
    \label{fig:framework}
\end{figure*}
\subsection{Decoupling Exploration and Exploitation}
Open-world tasks differ from traditional reasoning and decision-making tasks. Traditional reasoning~\cite{huang2022towards,sun2023survey} or decision-making~\cite{yang2023foundation} tasks typically present all relevant information at once, requiring the agent to deduce and make a plan based on the provided information, such as mathematical calculations or logical reasoning problems. Conversely, in open-world tasks, only the task description is initially specified. In this context, the agent must continually interact with the environment to obtain supporting information, comprising the exploration and exploitation steps.

Let $E$ and $T$ represent the environment and the task, $\Theta$ denote the LLM, and $P$ denote the prompt. The action space of the agent is defined as $\mathcal{A}=\mathcal{A}_e \cup \mathcal{A}_t$, where $\mathcal{A}_e$ and $\mathcal{A}_t$ represent the action set of exploration and exploitation, respectively. Exploration and exploitation are denoted by the functions $explore(\cdot)$ and $exploit(\cdot)$. The information given by the environment in the $i$-th step is denoted as $F_i$. Regarding existing methods such as ReAct~\cite{yao2022react} where exploration and exploitation steps are embedded in reasoning, the action taken at step $i$ is represented as follows:
\begin{equation*}
\small
\centering
a_i= reason(E, T, s_{i-1}; \Theta, P, K=\cup_{j<i}\{F_j\}) \in \mathcal{A}_e\cup \mathcal{A}_{t}.
\end{equation*}
where $reasion(\cdot)$ denotes the mix of explore and exploit.

\begin{algorithm}[t]
    \small
    \KwIn{Knowledge triplets set $K$.}
    \KwOut{Knowledge graph $G$.}
    {
        % \tcp{Exploration with weak LLM agent.}
        Entity set $E \gets \{\}$\;
        Relation set $R \gets \{\}$\;
        Adjacency matrix $M$\;
        \For{$x \in K$}
        {
            $h, r, t \gets x$\;
            $E \gets E \cup \{h, t\}$;  $R \gets R \cup \{r\}$; $M[h][t] \gets r$\;
        }
        $G.E \gets E$; $G.R \gets R$; $G.M \gets M$\;
    }
\caption{Graph construction algorithm.}
\label{alg:construct_graph}
\end{algorithm}

% \vspace{-0.3cm}
\begin{algorithm}[htb]
    \small
    \KwIn{Knowledge graph $G$, Task $T$, LLM $\Theta$.}
    \KwOut{Triplets set $K$.}
    {
        
        Task-related entity set $E\gets extract(G.E, T; \Theta)$\;
        $K \gets \{\}$\;
        \For{$e_i \in E$}
        {
            \For{$e_j \in E\setminus\{e_i\}$}
            {
                $r\gets G.M[e_i][e_j]$\;
                \If{$r\neq \text{empty}$}
                {
                    $K \gets K\cup \left\{\big(e_i, r, e_j\big)\right\}$\;
                }
            }
        }
    }
\caption{Triplet retrieval algorithm.}
\label{alg:retrieve_triplet}
\end{algorithm}
% \vspace{-0.3cm}

Within this paradigm, the knowledge $K$ utilized is solely the limited information about the environment obtained through partial observations. Particularly, greedy decisions are taken in the initial steps when the agent possesses limited awareness of the environment. For instance, in a task such as ``\textit{cleaning some apples with soap}" and the agent's initial location is the hall. The actual locations of the apple and soap are in the drawer of the table in the hall and on the sink in the kitchen, respectively. The lack of environmental knowledge may lead the agent to be misled by the world knowledge of the LLM, going to the kitchen to find the apple. Consequently, substantial efforts traversing every corner of the kitchen are wasted, resulting in suboptimal plans and even failures due to trapping in the loop. Therefore, we investigate the strategy to decouple exploration and exploitation, formalized as follows:
\begin{equation*}
\small
a_i = \left\{
\begin{aligned}
    & explore(E, T, s_{i-1}; \Theta, P_e) \in \mathcal{A}_e,\; i<N_e; \\
    & exploit(E, T, s_{i-1}; \Theta, P_t, K=\cup_{j\le N}\{F_j\}) \in \mathcal{A}_t,  i\ge N_e.
\end{aligned}
\right.
\end{equation*}
where $P_e$ and $P_t$ represent the prompts of the exploration and exploitation task, respectively. $N_e$ is the maximum number of steps of exploration, which could also be determined by the agent, such as terminating the exploration automatically when it thinks the obtained information is sufficient.

Different from the previous methods, our method places the whole exploration phase before exploitation explicitly, as opposed to the alternation of exploration and exploitation. In this manner, the agent has extensively explored the environment, acquiring global environmental prior knowledge denoted as \(K=\cup_{j\le N}\{F_j\}\). Exploitation with global knowledge benefits the effectiveness and efficiency of the solutions, which is empirically validated in our experiments.

However, two subsequent issues exist following the decoupling approach. Firstly, the information obtained from environmental feedback is huge due to the extensive exploration, including a lot of task-irrelevant information. Secondly, the extensive exploration contributes to increased resource consumption, such as token usage. Therefore, we demand an efficient mechanism for information transfer between exploration and exploitation and a cost-efficient exploration-exploitation strategy. We address the two issues in the subsequent parts of this section.
% 开放世界中探索和利用的概念以及关系，形式化描述
% 探索和利用结合 / 解耦的区别
% \subsubsection{Relation with Encoder-Decoder arch.}

\subsection{Knowledge Compression and Retrieval}

Real-world textual information exhibits inherent sparsity, characterized by long sentences consisting of plenty of non-informative conjunctions and adjectives. Environmental feedback in open-world tasks manifests as such text, where the cumulative extensive exploration yield long and unstructured textual information, demonstrating serve sparsity. Considering the limited context window of the LLM and the expensive cost of token usage, it is necessary to compress the sparse information. Leveraging a knowledge graph (KG) to store information has proved advantageous in enhancing information density and leveraging domain-specific knowledge in existing works~\cite{pan2024unifying}. 

Consequently, benefiting from the superiority of LLM in relation-extraction tasks~\cite{wadhwa2023revisiting}, we extract the knowledge from the received feedback to form an environmental knowledge graph. Specifically, the LLM extracts knowledge triplets from the environmental feedback after each exploration step, updating them into the knowledge graph. For example, as for the search result given by Wikipedia ``\textit{Since 2005 Wendy Schaal has primarily worked in voice acting, most notably voicing Francine Smith in the animated comedy television series American Dad!}", knowledge triplets are extracted as $\langle$\textit{Wendy Schaal}, \textit{voice for}, \textit{Francine Smith}$\rangle$ and $\langle$\textit{Francine Smith}, \textit{character in}, \textit{American Dad!}$\rangle$. Notably, the environmental knowledge graph we obtained is task-relevant, serving as a memory like the Random Access Memory(RAM). Actually, a worldwide knowledge graph could be leveraged and continually in our method, serving as a general memory. We leave it for further work.

\begin{algorithm}[thb]
% \centering
    \small
    \KwIn{Environment $E$, Task $T$, Initial state $s_0$, Weak LLM $\Theta_{w}$, Strong LLM $\Theta_{s}$, Exploration prompt $P_e$, Exploitation prompt $P_t$, Limit of steps $N_e,N_t$.}
    \KwOut{Plan $p$.}
    {
        \tcp{Exploration with weak LLM agent.}
        $K \gets \{\}$; $i \gets 0$; $s^e_i \gets s_0$\;
        \For{$i<N_e$}
        {
            $a^e_i \gets explore(E, T, s^e_i; \Theta_{w}, P_e)$\;
            $s^e_i, F_i \gets step(E, s^e_i, a^e_i)$\;
            $K^{'} \gets extract(F_i; \Theta_{w})$\;
            $K \gets K\cup K^{'}$\;
            $i \gets i+1$\;
        }

        $G_K \gets construct\_graph(K)$ \tcp*{Alg~\ref{alg:construct_graph}}
        \tcp{Exploitation with strong LLM agent.}
        $i \gets 0$; $s^t_i \gets s_0$; $p \gets []$\;
        $\tilde{K} \gets retrieve\_triplets(G_K, T; \Theta_w)$ \tcp*{Alg~\ref{alg:retrieve_triplet}}
        \For{$i<N_t$ and $F_i\neq \text{Completed}$}
        {
            $a^t_i \gets exploit(E, T, s^t_i; \Theta_{s}, P_t, \tilde{K})$\;
            $s^t_i, F_i \gets step(E, s^t_i, a^t_i)$\;
            $i \gets i+1$\;
            $p \gets p+[a^t_i]$\;
        }
    }
\caption{WESE algorithm.}
\label{alg:wese}
\end{algorithm}

Nevertheless, it is imperative to acknowledge that not all information in the knowledge graph proves useful. The introduction of task-irrelevant information has the potential to lead the hallucination phenomena of LLM, such as the confusion of entity and relation. 
For example, giving the triplet $\langle$\textit{Bob}, \textit{favorite fruit}, \textit{apple}$\rangle$ and the question is ``\textit{What's the favorite fruit of Bill?}", the LLM would confuse the relation and answer with \textit{apple}.
Benefiting from the graph structure, we adopt a one-hop retrieval method to extract task-related information easily, illustrated in Algorithm~\ref{alg:retrieve_triplet}. 
Concretely, we initiate the process by extracting involved entities from the task description with LLM. Subsequently, we perform a one-hop retrieval on the graph to obtain the neighbors of these entities. The retrieved knowledge triplets are then injected into the prompt, serving as task-relevant knowledge during the exploitation phase, thereby assisting the LLM in task-solving.

\subsection{Weak Exploration to Strong Exploitation}

\begin{table*}[th]
    \small
    \centering
    \caption{Results on ALFWorld(134 tasks). \textbf{SR} and \textbf{AS} are abbreviations for success rate and average steps of successful tasks, respectively. \textbf{SESE} represents the variant of WESE---Strong Exploration to Strong Exploitation. The \textit{Imp} represents the relative improvements compared to base methods, i.e. Act and ReAct. The \textbf{bold} and \underline{underline} represent the best and the second best for the same base method.}
    \vspace{-0.3cm}
    \begin{tabular}{l|rr|rr|rrrr}
        \toprule
        Performance & \multicolumn{2}{c|}{Effectiveness} & \multicolumn{2}{c|}{Efficiency} &  \multicolumn{4}{c}{Cost} \\ \midrule
        Method & SR$\uparrow$ & \textit{Imp}(\%)  & AS$\downarrow$ & \textit{Imp}(\%) &  Prompt$\downarrow$ & Completion$\downarrow$ & {Expense}(\$)$\downarrow$ & \textit{Imp}(\%) \\ \midrule
        Act & 0.43 & 0.00 & 10.83 & 0.00 & 4,908,548 & 21,243 & \underline{98.60} & 0.00\\
        Act-\textbf{WESE} & \underline{0.63} & +46.51 & \underline{7.54} & +30.38 & 3,746,290 & 19,562 & \textbf{75.32} & +23.61 \\
        Act-\textbf{SESE} & \textbf{0.67} & +55.81 & \textbf{6.73} & +37.86 & 7,259,508 & 75,153 & 146.69 & -48.77 \\
        % Reflexion & \\ 
        \midrule
        ReAct & 0.57 & 0.00 &  16.64 & 0.00 & 7,565,676 & 43,250 & \underline{152.18}  & 0.00 \\
        ReAct-\textbf{WESE} & \underline{0.72} & +26.32 & \underline{13.69} & +17.73 & 5,032,374 & 41,004 & \textbf{101.47} & +33.32  \\
        ReAct-\textbf{SESE} & \textbf{0.75} & +31.58 & \textbf{12.41} & +25.42 & 8,996,182 & 97,286 & 181.87 & -19.51  \\
        % Reflexion-WESE & \\ 
        \bottomrule
    \end{tabular}
    \label{tab:alfworld}
\end{table*}

\begin{table*}[th]
    \small
    \centering
    \caption{Results on ScienceWorld(296 tasks). \textbf{TR}, \textbf{AR} and \textbf{AS} are abbreviations for total reward, average reward and average steps to get positive reward, respectively. Other symbols are consistent with Table~\ref{tab:alfworld}.}
    \vspace{-0.3cm}
    \begin{tabular}{l|rrr|rr|rrrr}
        \toprule
        Performance & \multicolumn{3}{c|}{Effectiveness} & \multicolumn{2}{c|}{Efficiency} &  \multicolumn{4}{c}{Cost} \\ \midrule
        Method & TR$\uparrow$ & AR$\uparrow$ & \textit{Imp}(\%) & AS$\downarrow$ & \textit{Imp}(\%) &  Prompt$\downarrow$ & Completion$\downarrow$ & {Expense}(\$)$\downarrow$ & \textit{Imp}(\%)\\ \midrule
        Act & 4908 & 16.58 & 0.00 & 18.00  & 0.00 & 13,554,960 & 55,817 & \underline{272.22}  & 0.00 \\
        Act-\textbf{WESE} & \underline{5198} & \underline{17.56} & 5.91 & \underline{15.68}  & +12.91 & 13,491,043 & 65,952 & \textbf{271.14}  & +0.40 \\
        Act-\textbf{SESE} & \textbf{5249} & \textbf{17.73 }& 6.94 & \textbf{15.39 } & +14.49 & 36,424,190 & 165,568 & 731.80  & -168.83 \\
        \midrule
        ReAct & 4454 & 15.05 & 0.00 & 20.00  & 0.00 & 17,716,698 & 84,724 & \underline{356.03}  & 0.00 \\
        ReAct-\textbf{WESE} & \textbf{5317} & \textbf{17.96} & 19.34 & \underline{19.65}  & +1.77 & 16,310,632 & 80,851 & \textbf{327.83}  & +7.92 \\
        ReAct-\textbf{WESE} & \underline{5053} & \underline{17.07} & 13.42 & \textbf{19.02}  & +4.92 & 40,293,571 & 196,338 & 809.80  & -127.45 \\
        % Reflexion-WESE & \\ 
        \bottomrule
    \end{tabular}
    \label{tab:sciworld}
\end{table*}

Acquiring more comprehensive global information about the environment demands a considerable resource cost in the exploration process. However, compared to exploitation, exploration exhibits lower complexity, requiring less reasoning and induction.
Concretely, exploration operations exhibit low requirements for the logic and coherence of actions, emphasizing actions pertaining to environmental observation.
For example, the exploration actions mainly consist of several simple actions on decision-making benchmarks, such as ``\textit{go to} [\textsc{room}]", ``look around", et al, while exploitation involves a series of coherent operations like (\textit{go to sink/stove, put the bowl in/on the sink/stove, activate the sink/stove, wait, deactivate the sink/stove}).
Therefore, we propose to use a weaker agent for the exploration to mitigate resource consumption, namely the \textbf{weak exploration}. From the perspective of the LLM agent, a weaker agent represents substituting the underlying LLM for exploration with a weaker LLM, i.e. an LLM with fewer parameters, thereby reducing costs. In our experiments, we compare performance between strong exploration and weak exploration. Our findings reveal that a weaker exploration has a negligible impact on the final success rate, yet it significantly lowers costs.
% why weaker LLM cost-saving, details

The framework of WESE is illustrated in Figure~\ref{fig:framework}. There are three key components in the framework: a weak LLM agent, a strong LLM agent, and a KG-based memory. The whole process consists of the weak exploration (left) and the strong exploitation (right). Meanwhile, we offer an algorithmic pseudo-code in Algorithm~\ref{alg:wese}. First, a weak LLM agent is employed to explore the interactive environment to obtain information in line 1 to 7. Then those knowledge triplets are organized as a knowledge graph $G_K$ in line 8, as illustrated in Algorithm~\ref{alg:construct_graph}. Further, the involved entities are extracted from the task with a LLM and the relevant triplets are retrieved from the graph in line 10. Retrieved knowledge is leveraged for exploitation in line 12, serving as the prior knowledge.

\section{Experiments}

We employ two categories of interactive open-world tasks as benchmarks: decision-making and question-answering, where each task requires multi-step interactions with the environment. We evaluate our methods from three perspectives: effectiveness, efficiency and cost, representing whether the agent can complete the tasks, how many steps the agent would take to finish the task, and the expenses for the agent to complete the task, respectively.

\subsection{Decision Making Tasks}
We begin with the open-world decision-making tasks, where environments are based on a text-based simulator. The tasks are about the household, where the agent needs to explore various rooms and take operations on several objects.

\subsubsection{ALFWorld}
ALFWorld~\cite{shridhar2020alfworld} is a synthetic text-based simulated interactive environment. It comprises six types of tasks where agents need to interact with the environment to generate a series of actions to solve household tasks. For example, in the task ``clean some knife and put it in countertop", the ideal solution involves actions such as (go to countertop 2, take knife 1, go to sinkbasin 2, clean knife 1, put knife 1 on countertop 2). These tasks vary in difficulty, with challenging tasks encompassing over 50 locations and requiring more than 50-step actions, posing challenges for both the exploration and exploitation processes.

\begin{figure}[thb]
    \centering
    % \vspace{-10pt}
    \subfigure[Relative improvements for Act-based methods.]{
        \includegraphics[width=0.85\columnwidth]{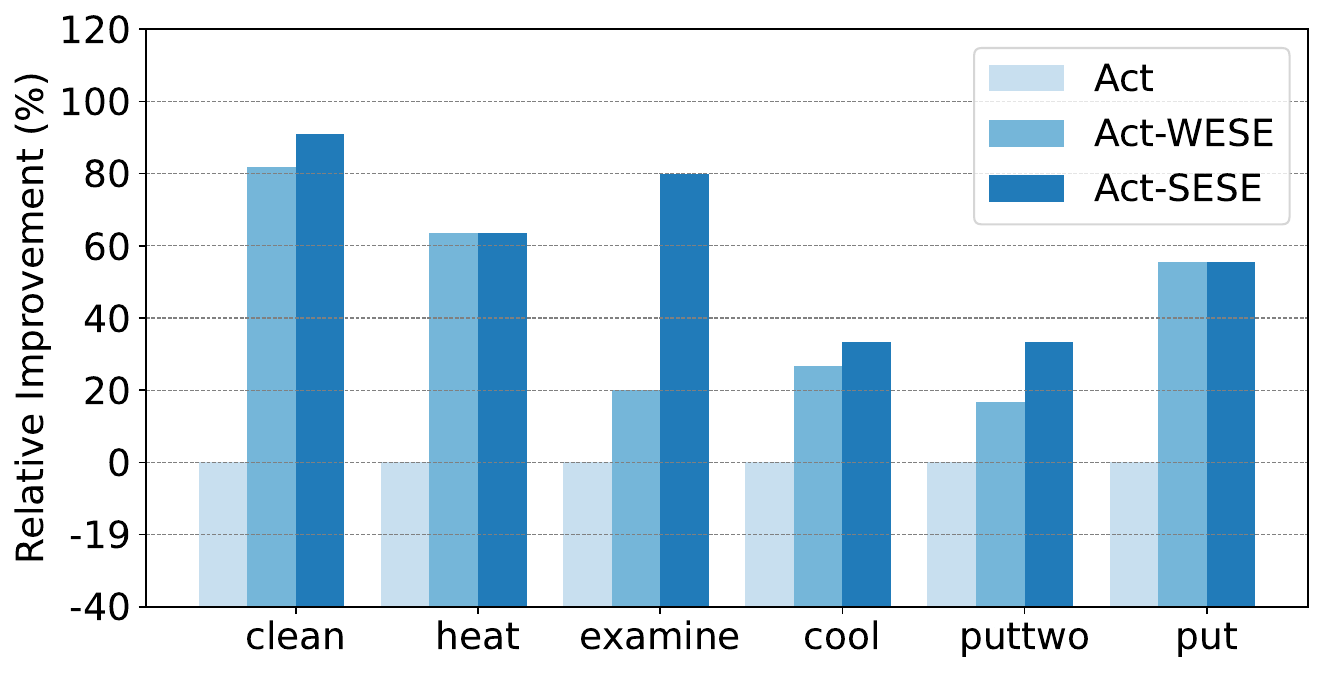}
        \label{fig:alf_act}
        % \vspace{-20pt}
    }
    % \vspace{-10pt}
    \subfigure[Relative improvements for ReAct-based methods.]{
        \includegraphics[width=0.85\columnwidth]{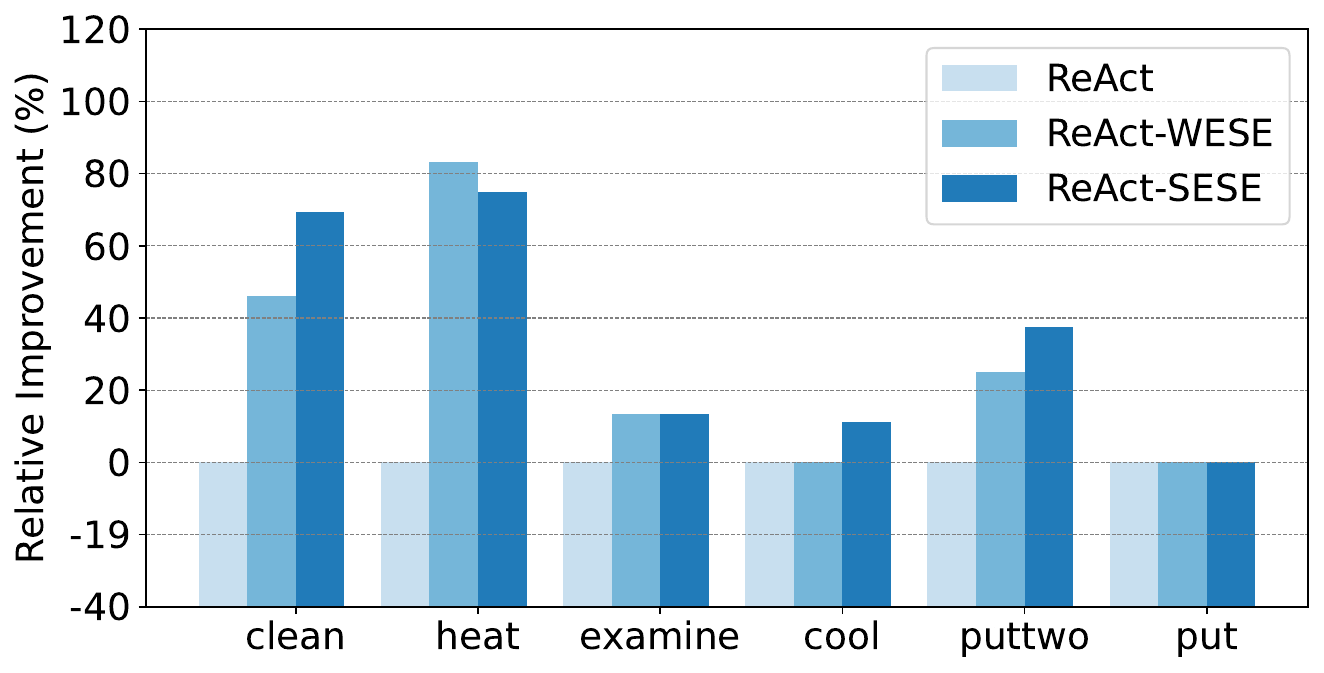}
        \label{fig:alf_react}
    }
    \vspace{-0.3cm}
    \caption{Relative improvements in success rate over various types of tasks on ALFWorld. The left tasks are more complicated.}
    \label{fig:alf_tasks}
    % \vspace{-0.3cm}
\end{figure}

To validate the effectiveness of WESE, we adopt Act~\cite{yao2022react} and ReAct~\cite{yao2022react} as baselines. Act leverages the idea of CoT, providing LLMs with few-shot interactive examples. ReAct, building upon Act, introduces an extra ``\textsc{thought}'' step where LLMs can choose to explicitly output their thought about the current state or generate action. In WESE, we initially use a Weak LLM for exploration to acquire task-relevant knowledge. We then leverage the obtained knowledge to solve problems with two base methods. We employ Llama-2-7B~\cite{touvron2023llama} as the weak LLM and text-davinci-003 (with probably more than 175 billion parameters) developed by OpenAI~\footnote{\url{https://platform.openai.com/}} as the strong LLM. The limits of steps $N_e, N_t$ are both set to 50. Our evaluation focuses on success rates, average steps to complete tasks, and the cost of OpenAI API tokens as three key metrics. Additionally, we introduce a variant of WESE---Strong Exploration to Strong Exploitation (\textbf{SESE}), where the weak LLM in the exploration process is replaced with the strong LLM, to verify the effectiveness of the decoupling strategy and examine the impact of LLM strength on exploration quality.

\subsubsection{ScienceWorld}
Similar to ALFWorld, ScienceWorld~\cite{wang2022scienceworld} is an interactive household environment as well. However, the tasks in ScienceWorld are more challenging, involving scientific experiments such as boiling and creating a new color by mixing primary colors. The environment is more complex, comprising ten distinct rooms, each with different furnishings, and not each pair of rooms is connected.

We conduct experiments on eight types of tasks within ScienceWorld, choosing about 30 instances for each task due to a limited budget. Unlike ALFWorld where the agent can get a reward of 1 only when the task is completed, the agent in ScienceWorld receives partial rewards upon completing crucial steps, with the total reward reaching 100. Given the challenging nature of the tasks, achieving a full reward of 100 is rare. Therefore, we utilize the number of steps taken by the agent until it first obtains a positive reward as the metric for efficiency. Other settings are consistent with ALFWorld.

% \subsubsection{Baselines}

\subsubsection{Results}
The results on ALFWorld and ScienceWorld are shown in Table~\ref{tab:alfworld} and Table~\ref{tab:sciworld}, respectively. We conclude several findings based on the results.
Consistent with results reported in ReAct, ReAct outperforms Act on two benchmarks, showing the superiority of the ``\textsc{thought}'' step. However, this additional step leads to a longer action sequence, resulting in an average relative 32.38\% increase in average steps. 
Decoupling of exploration and exploitation demonstrates advantages in effectiveness and efficiency, resulting in SESE outperforming baselines significantly with average relative 26.94\% and 20.67\% improvements in terms of success rate (average reward) and average steps. However, the cost of SESE increases a lot due to the introduction of extensive strong exploration, showing an average relative 91.14\% increase over baselines. 

WESE shows a better balance between effectiveness, efficiency, and cost, which saves 53.83\% of costs with only relative 1.43\% and 6.89\% degradations in effectiveness and efficiency compared with SESE. In WESE, the weak LLM agent undertakes the exploration process, resulting in cost savings for extensive exploration. Besides, benefiting from the related triplets extracted from the explored KG, the strong LLM agent only needs to focus on exploitation, further decreasing the number of steps, evidenced by the decreased completion tokens and average steps.

We further investigate the improvements of WESE on various types of tasks, shown in Figure~\ref{fig:alf_tasks}. Both WESE and SESE show improvements over almost all types of tasks, further indicating the effectiveness of the decoupling strategy. In addition, the improvements in ``\textit{clean}" and ``\textit{heat}" tasks are greater than other tasks. The reason lies in that the two tasks involved more complicated exploitation compared with ``\textit{put}", where the agents need to find the object first and then clean or heat it instead of just moving it to another place. The result demonstrates extensive exploration benefits more for complex tasks.

\subsection{Question Answering Tasks}
We also validate our WESE on two open-world interactive question-answering benchmarks, i.e., HotPotQA and FEVER. Different from traditional question-answering tasks where supporting sentences are given, those tasks provide the question only and require the agent to search information on the web step by step to give the final answer.

\begin{table*}[th]
    \small
    \centering
    \caption{Results on HotPotQA(500 tasks). \textbf{SR} and \textbf{AS} are abbreviations for success rate and average steps of successful tasks, respectively. \textbf{SESE} represents the variant of WESE---Strong Exploration to Strong Exploitation. The \textit{Imp} represents the relative improvements compared to base methods, i.e. Act and ReAct. The \textbf{bold} and \underline{underline} represent the best and the second best for the same base method.}
    \vspace{-0.3cm}
    \begin{tabular}{l|rr|rr|rrrr}
        \toprule
        Performance & \multicolumn{2}{c|}{Effectiveness} & \multicolumn{2}{c|}{Efficiency} &  \multicolumn{4}{c}{Cost} \\ \midrule
        Method & SR$\uparrow$ & \textit{Imp}(\%) & AS$\downarrow$ & \textit{Imp}(\%) &  Prompt$\downarrow$ & Completion$\downarrow$ & {Expense}(\$)$\downarrow$ & \textit{Imp}(\%)\\ \midrule
        CoT & 0.318 & N/A & {1.00} & N/A & 261,347 & 25,382 & {5.73} & N/A\\
        \midrule
        Act & 0.296 & 0.00 & 3.53 & 0.00 & 2,390,041 & 14,236 & \underline{48.09} & 0.00 \\
        Act-\textbf{WESE} & \underline{0.353} & +19.26  & \underline{2.69} & +23.80 & 2,307,421 & 13,973 & \textbf{46.42} & +3.45\\
        Act-\textbf{SESE} & \textbf{0.361} & +21.96 & \textbf{2.58} & +26.91 & 7,522,826 & 27,1551 & 155.89 & -224.18\\
        % Reflexion & \\ 
        \midrule
        ReAct & 0.342 & 0.00 & 3.17 & 0.00 & 3,234,876 & 65,306 & \underline{66.00} & 0.00 \\
        ReAct-\textbf{WESE} & \underline{0.394} &  +15.20 & \underline{2.29} & +27.76 & 2,574,401 & 67,908 & \textbf{52.85} & +19.93\\
        ReAct-\textbf{SESE} & \textbf{0.416} & +21.64 & \textbf{2.11} & +33.44 & 7,338,590 & 323,401 & 153.24 & -132.17 \\
        % Reflexion-WESE & \\ 
        \bottomrule
    \end{tabular}
    \label{tab:HotPotQA}
\end{table*}

\begin{table*}[th]
    \small
    \centering
    \caption{Results on FEVER(500 tasks). The meanings of abbreviations and symbols are consistent with Table~\ref{tab:HotPotQA}.}
    \vspace{-0.3cm}
    \begin{tabular}{l|rr|rr|rrrr}
        \toprule
        Performance & \multicolumn{2}{c|}{Effectiveness} & \multicolumn{2}{c|}{Efficiency} &  \multicolumn{4}{c}{Cost} \\ \midrule
        Method & SR$\uparrow$ & \textit{Imp}(\%) & AS$\downarrow$ & \textit{Imp}(\%) &  Prompt$\downarrow$ & Completion$\downarrow$ & {Expense}(\$)$\downarrow$ & \textit{Imp}(\%)\\ \midrule
        CoT & 0.61 & N/A & 1.00 & N/A & 100,387 & 11,942 & 2.25 & N/A \\
        \midrule
        Act & 0.56 & 0.00 & 2.16 & 0.00 & 723,646 & 6,980 & \underline{14.61} & 0.00 \\
        Act-\textbf{WESE} & \underline{0.62} & +10.71 & \underline{1.58} & +26.66 & 723,867 & 5,937 & \textbf{14.60} & +0.11\\
        Act-\textbf{SESE} & \textbf{0.64} & +14.29 & \textbf{1.57} & +27.34 & 2,822,189 & 122,543 & 60.89 & -316.73 \\
        % Reflexion & \\ 
        \midrule
        ReAct & 0.63 & 0.00 & 2.18 & 0.00 & 1,074,080 & 36,040 & \underline{22.20} & 0.00  \\
        ReAct-\textbf{WESE} & \underline{0.68} & +7.26 & \underline{1.62} & +25.96 & 918,905 & 29,895 & \textbf{18.98} & +14.53  \\
        ReAct-\textbf{SESE} & \textbf{0.70} & +10.09 & \textbf{1.59} & +27.18 & 3,104,924 & 162,363 & 65.35 & -194.32  \\
        % Reflexion-WESE & \\ 
        \bottomrule
    \end{tabular}
    \label{tab:FEVER}
\end{table*}

\subsubsection{HotPotQA}
HotPotQA\cite{yang2018hotpotqa} is a question-answering dataset where each question is paired with supporting sentences from Wikipedia articles. In traditional QA tasks, the supporting sentences are given and the remained task is to reason. Referred in ReAct, we use the Wikipedia API with three types of actions to support interactive information retrieval: (1) \textsc{\textbf{search}}[\textsc{entity}], which searches the Wikipedia with the \textsc{entity} and returns the corresponding page if it exists, or suggests top-5 similar entities; (2) \textsc{\textbf{lookup}}[\textsc{keyword}], which looks up keyword in the page and returns the next sentence containing the \textsc{keyword}, simulating the Ctrl+F function in a web browser; (3) \textsc{\textbf{finish}}[\textsc{answer}], which answers the question with \textsc{answer}. Once the \textsc{answer} matches the ground truth, the environment would return reward 1. We sample 500 tasks from the development set.

We employ the CoT~\cite{wei2022chain}, Act and ReAct as baselines and empower Act and ReAct with WESE and SESE. Note that CoT is a one-step method that does not support interactive tasks, we inject the supporting sentences into the prompts and instruct the LLM to reason for the final answer without searching on the web. Also, WESE is not designed for such a purely reasoning method but for methods involving interactions with the environment. For Act and ReAct, we keep the settings consistent with the original paper. As there are probably lots of related triplets to the task-involved entities, we set the limit of retrieved triplets as 10 and the limits of steps $N_e, N_t$ as 8. The evaluation for effectiveness, efficiency and cost is consistent with the ALFWorld.

\subsubsection{FEVER}
FEVER~\cite{thorne2018fever} is a fact verification dataset, consisting of instances where each instance comprises a claim and a justification(\textsc{True} or \textsc{False} or \textsc{Not Clear}). We employ the Wikipedia API to construct an interactive environment consistent with that in HotPotQA. Other settings are kept consistent with HotPotQA, such as the number of retrieved triplets and the maximum steps.

\subsubsection{Results}
The results on HotPotQA and FEVER are shown in Table~\ref{tab:HotPotQA} and Table~\ref{tab:FEVER}, respectively. We can conclude several findings based on the results.
Similar to decision-making tasks, ReAct outperforms Act significantly due to the additional ``\textsc{thought}" step.
Also, methods equipped with WESE or SESE outperform baselines in both success rate and the number of taken actions, resulting in average relative improvements of 19.5\% and 28.0\%, respectively.
Especially, SESE methods surpass WESE slightly with average relative 3.5\% and 3.6\% improvements in terms of success rate and average steps, while increasing more than twice the expenses. This further demonstrates that the weak agent powered by Llama-2-7B is almost sufficient for the exploration task.

Different from decision-making tasks, question-answering tasks require fewer steps due to more information being returned with one search action. However, our WESE and SESE are still capable of reducing the number of steps, further showing the advantage of the explored knowledge.
As for the cost, the tokens increased in SESE are far more than those in decision-making tasks, which can be attributed to the long-textual feedback from Wikipedia.
\section{Conclusion}
In this paper, we introduce WESE, a cost-effective method that enhances LLM agents in open-world interactive tasks. We decouple the exploration and exploitation, employing two agents for the distinct processes. To empower the communication between the two processes, we introduce a knowledge graph-based memory to compress and structure the information obtained in exploration, where task-relevant information is extracted from the graph by a one-hop retrieval method. We then propose to leverage a weaker agent for the exploration process, forming a cost-effective manner with negligible performance degradation. Experimental results demonstrate the superiority of WESE in effectiveness, efficiency, and cost.

%% The file named.bst is a bibliography style file for BibTeX 0.99c
\newpage
\bibliographystyle{named}
\bibliography{myref}

\end{document}